\documentclass[10pt,twocolumn,letterpaper]{article}

\usepackage{iccv}              
\usepackage{makecell}
\usepackage{verbatim} 
\usepackage{bm}
\definecolor{iccvblue}{rgb}{0.21,0.49,0.74}
\usepackage[pagebackref,breaklinks,colorlinks,allcolors=iccvblue]{hyperref}
\usepackage{units}

\def\paperID{16} 
\def\confName{ICCV}
\def\confYear{2025}

\title{Improving Lightweight Weed Detection via Knowledge Distillation\thanks{First submitted version, accepted by the CVPPA at ICCV 2025.}}

\author{Ahmet O\u{g}uz Salt{\i}k, Max Voigt, Sourav Modak, Mike Beckworth, Anthony Stein \\\\
University of Hohenheim\\
Dept. of Artificial Intelligence in Agricultural Engineering \& Computational Science Hub\\
Stuttgart, Germany\\
{\tt\small \{ahmet.saltik, max.voigt, s.modak , mike.beckworth, anthony.stein\}@uni-hohenheim.de}
}

\begin{document}
\maketitle

\begin{abstract}
Weed detection is a critical component of precision agriculture, facilitating targeted herbicide application and reducing environmental impact. However, deploying accurate object detection models on resource-limited platforms remains challenging, particularly when differentiating visually similar weed species commonly encountered in plant phenotyping applications. In this work, we investigate Channel-wise Knowledge Distillation (CWD) and Masked Generative Distillation (MGD) to enhance the performance of lightweight models for real-time smart spraying systems. Utilizing YOLO11x as the teacher model and YOLO11n as both reference and student, both CWD and MGD effectively transfer knowledge from the teacher to the student model. Our experiments, conducted on a real-world dataset comprising sugar beet crops and four weed types (Cirsium, Convolvulus, Fallopia, and Echinochloa), consistently show increased AP50 across all classes. The distilled CWD student model achieves a notable improvement of 2.5\% and MGD achieves 1.9\% in mAP50 over the baseline without increasing model complexity. Additionally, we validate real-time deployment feasibility by evaluating the student YOLO11n model on Jetson Orin Nano and Raspberry Pi 5 embedded devices, performing five independent runs to evaluate performance stability across random seeds. These findings confirm CWD and MGD as an effective, efficient, and practical approach for improving deep learning-based weed detection accuracy in precision agriculture and plant phenotyping scenarios.
\end{abstract}

\section{Introduction}
\label{sec:intro}

Precision agriculture increasingly leverages computer vision to enable targeted interventions, reduce chemical inputs, and improve crop yields~\cite{modak2025exploring,allmendinger}. Among these tasks, weed detection is particularly critical: by accurately identifying and localizing weeds, smart sprayers can apply herbicides only where needed, minimizing environmental impact and operational costs~\cite{Milioto2018ICRA}. In practical applications, the detection of weeds frequently relies on lightweight Field Camera Units (FCUs) mounted on mobile platforms, which impose strict real-time processing demands on subsequent detection systems.

State-of-the-art object detectors such as YOLO (You Only Look Once) have demonstrated strong accuracy on weed identification benchmarks~\cite{Jocher_Ultralytics_YOLO_2023,Redmon2016YOLO}, but their performance typically degrades when pruned or quantized for edge hardware. This tension between model capacity and inference speed has driven the development of compact backbones such as MobileNets~\cite{Howard2017MobileNets}, ShuffleNets~\cite{shufflenet}, and recent YOLO “Nano” variants such as YOLO11~\cite{Jocher_Ultralytics_YOLO_2023}, which trade some accuracy for significant latency improvements on devices like the Jetson Orin Nano and Raspberry Pi 5.

Knowledge Distillation (KD) provides a systematic framework to bridge this gap by transferring the dark knowledge of a large “teacher” network into a small “student” model~\cite{hinton_KD}. Early KD approaches matched logits or intermediate feature activations (e.g.\ FitNets~\cite{Romero2015FitNets}), while more recent methods incorporate relational and attention-based losses to better capture inter-channel or spatial dependencies~\cite{Zagoruyko2017Attention, Park2019Relational}. For dense‐prediction tasks, Channel‐wise Knowledge Distillation (CWD) has proven particularly effective: it treats each feature‐map channel as a spatial probability distribution and minimizes the Kullback–Leibler divergence between teacher and student, thereby aligning semantic attention patterns without heavy computational overhead~\cite{Shu_2021_ICCV}. Complementing this, Masked Generative Distillation (MGD) randomly masks portions of the student’s feature maps and forces the network to reconstruct the corresponding teacher features, yielding richer representations and consistently boosting performance across object detection and segmentation benchmarks~\cite{yang2022masked}.

We seek to answer the research question, how effectively do Channel‐wise Knowledge Distillation (CWD) and Masked Generative Distillation (MGD) transfer knowledge from a high‐capacity YOLO11x teacher to a lightweight YOLO11n student for real‐time weed detection, and how do the key hyperparameters, temperature $\tau$ for CWD and logit weight $\alpha$ for MGD, impact detection performance?

Accordingly, the main contributions of this work are three-fold:

\begin{itemize}
  \item We demonstrate a 2.5\% absolute improvement in mAP$_{50}$ on a real‐world sugar beet dataset via CWD, without increasing the student’s model complexity.
  \item We conduct a systematic study of the CWD distillation softmax temperature \(T\in\{1,2,3,4\}\) and MGD distillation logit-weight \(alpha\in\{2\cdot10^{-5}, 4\cdot10^{-5}, 6\cdot10^{-5}, 8\cdot10^{-5}\}\).
  \item We validate the distilled student’s performance and real‐time capabilities through five independent deployment trials on Jetson Orin Nano and Raspberry Pi 5 platforms, confirming its suitability for in‐field precision‐agriculture applications.


\end{itemize}

\section{Related Work}
\label{sec:related}

\subsection{Knowledge Distillation in Object Detection}
Knowledge Distillation (KD) transfers “dark knowledge” from a high-capacity teacher to a compact student, originally via softened logits matching~\cite{hinton_KD}. FitNets introduced intermediate feature supervision for classification tasks~\cite{Romero2015FitNets}, and subsequent works proposed relational and attention-based losses to capture richer inter-feature dependencies~\cite{Park2019Relational,Zagoruyko2017Attention}. In object detection, Channel-wise Knowledge Distillation (CWD) aligns per-channel spatial distributions to better transfer semantic attention, achieving SOTA gains on dense-prediction benchmarks~\cite{Shu_2021_ICCV}. More recently, Focal and Global Distillation (FGD) further decomposes feature maps into foreground-focused and global relational components, boosting mAP on COCO~\cite{Yang_2022_CVPR}.

\subsection{Knowledge Distillation in Agriculture}
Deploying KD in precision agriculture has shown substantial gains with minimal overhead. Jin et al. distilled classification networks for turfgrass weed detection~\cite{Jin2024PestManag}. Castellano et al. applied KD to UAV-based weed segmentation, achieving efficient aerial monitoring~\cite{Castellano_KD}. Zhou et al. combined feature-map and logit distillation in YOLOv5 for broadleaf weed detection, reporting significant mAP improvements~\cite{Zhou2023TASA}. Guo et al. presented YOLOv8n-DT for rice-field weed detection within an automated spraying pipeline~\cite{Guo2023YOLOv8nDT}. Ensemble-based KD, which fuses knowledge from multiple teachers, has also proven effective for crop–weed segmentation~\cite{Asad2024EnsembleKD}.

\subsection{Real-time Lightweight Models in Agriculture}
Edge deployments require detectors that balance accuracy and speed. Narayana \& Ramana demonstrated YOLOv7 for early-stage weed detection at real-time frame rates on embedded GPUs~\cite{Narayana2023YOLOv7Weed}. Sonawane et al. used Neural Architecture Search to design YOLO-NAS, optimizing accuracy–efficiency trade-offs for edge devices~\cite{Patel2024YOLONAS}. Gazzard et al. validated YOLOv8 for blackgrass classification on Jetson Nano, highlighting robustness under varying field conditions~\cite{Gazzard2024WeedScout}. Dang et al. benchmarked different YOLO versions across diverse weed species and lighting scenarios, confirming its practical suitability for precision spraying~\cite{Dang2024YOLOBenchmark}. Saltik et al. conducted a comparative analysis of YOLOv9, YOLOv10, and RT-DETR for real-time weed detection across multiple hardware configurations, elucidating the trade-offs between inference latency and detection accuracy~\cite{saltik2024comparative}. Herterich et al.~\cite{HERTERICH2025110608,Herterich_nas} presented a systematic experimental evaluation of lightweight object‐detection models deployed on NPUs, and also examining the impact of diverse backbone architectures and neural‐architecture‐search strategies.

\section{Material and Methods}
\label{sec:material_and_methods}

\subsection{Dataset}
\begin{figure}[!t]
  \centering
  %
  \begin{subfigure}[t]{0.49\linewidth}
    \includegraphics[width=\linewidth]{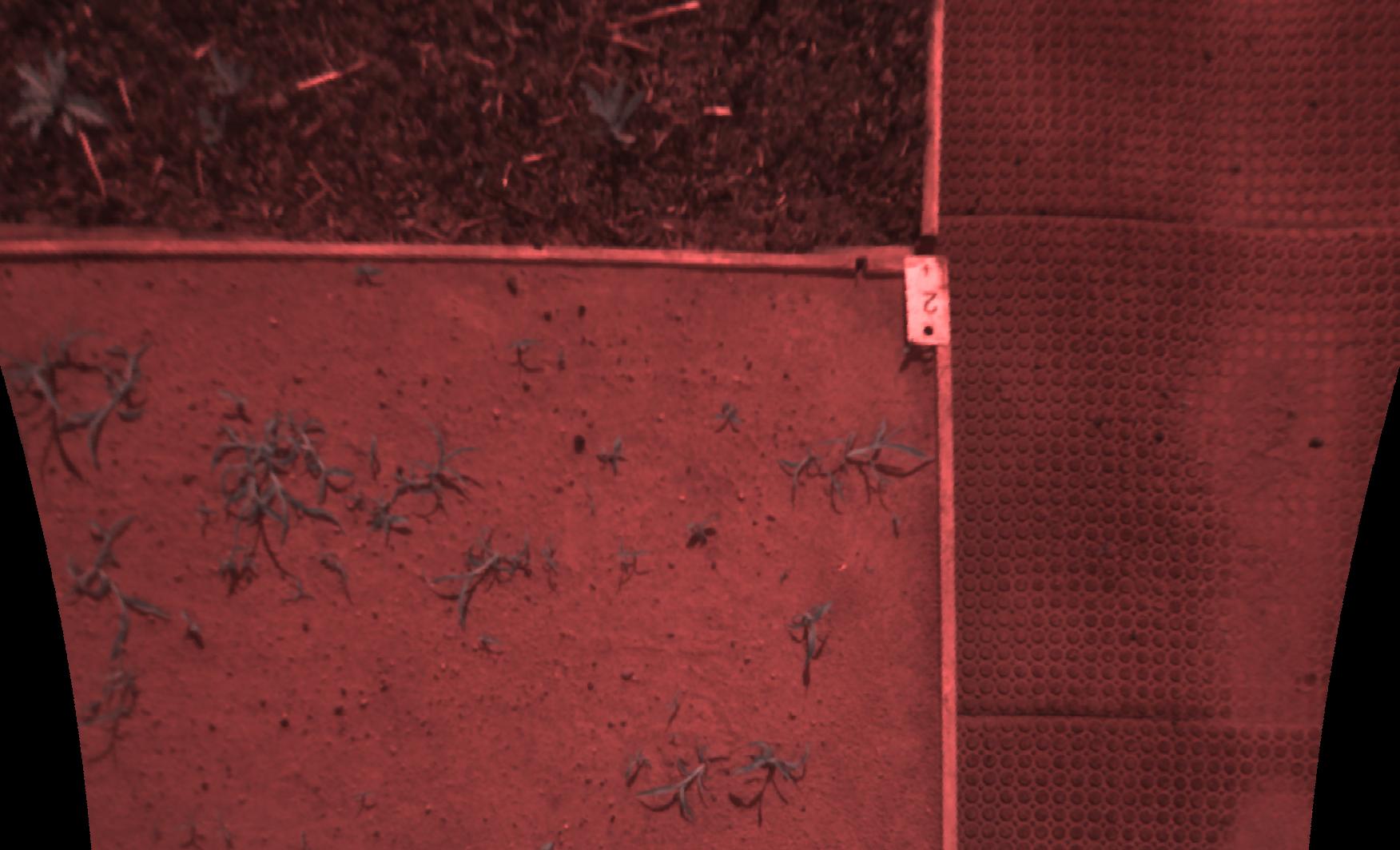}\\[0.50em]
    \includegraphics[width=\linewidth]{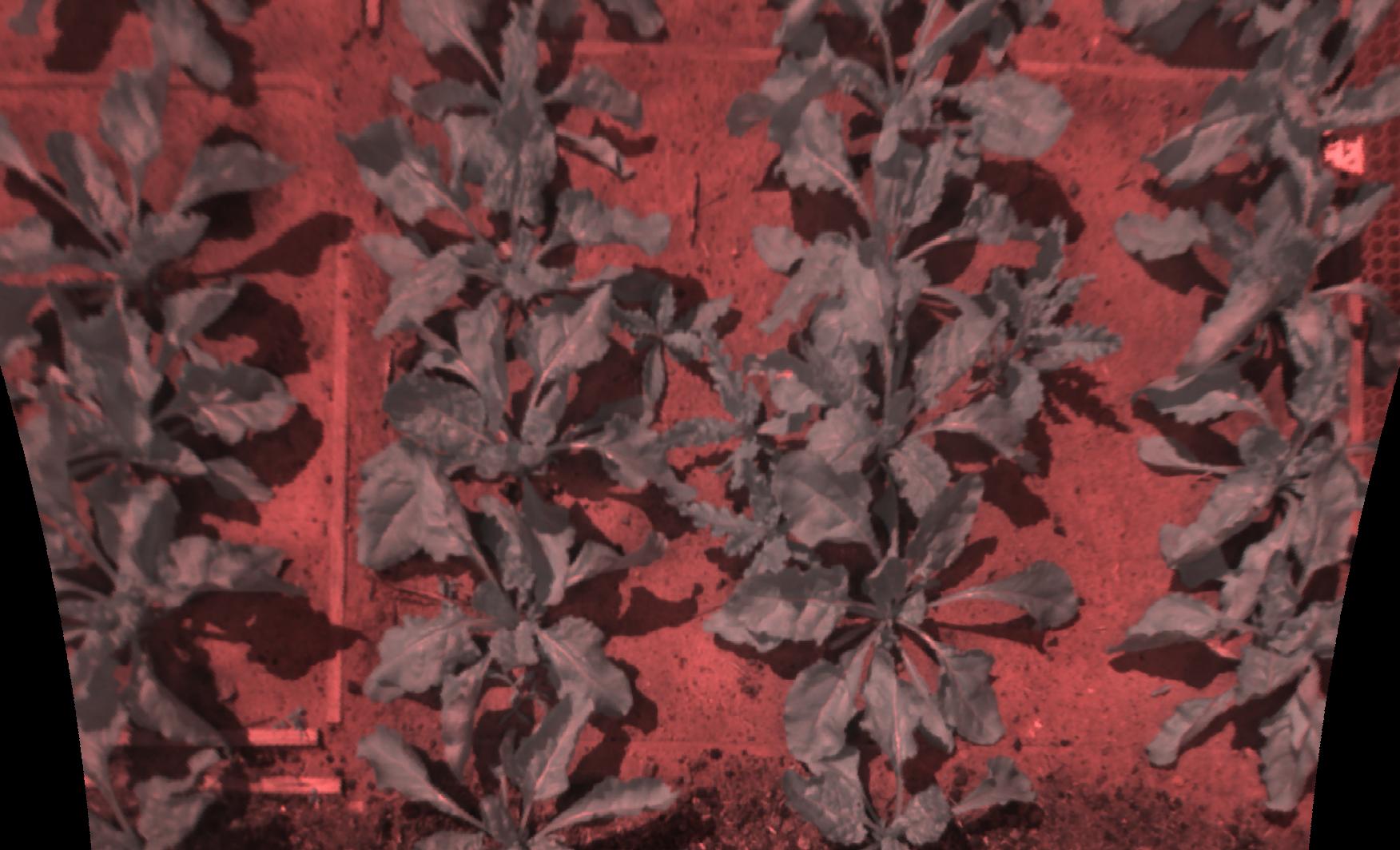}
    \caption{Captured images}
    \label{fig:captured}
  \end{subfigure}%
  \hfill
  %
  \begin{subfigure}[t]{0.49\linewidth}
    \includegraphics[width=\linewidth]{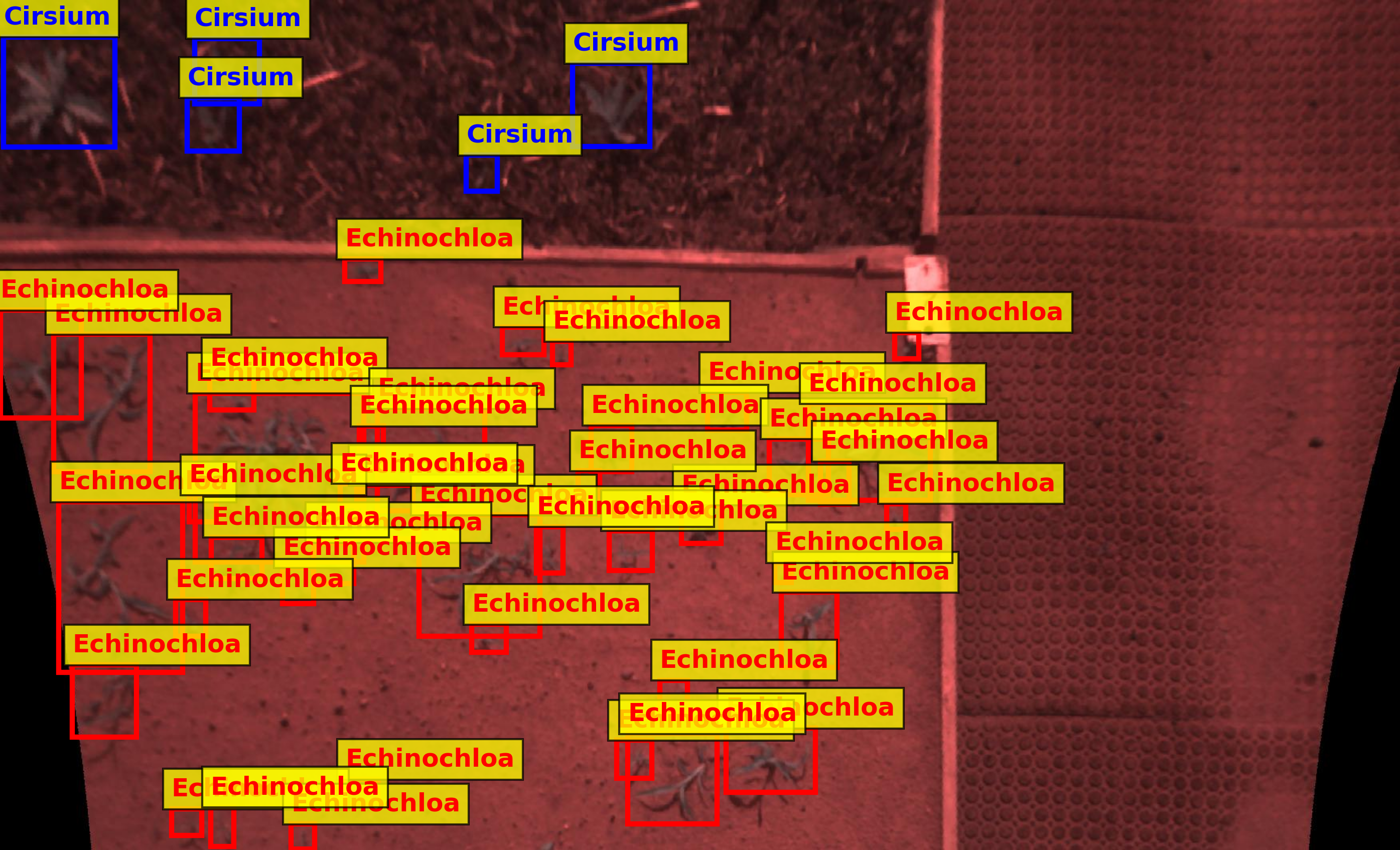}\\[0.25em]
    \includegraphics[width=\linewidth]{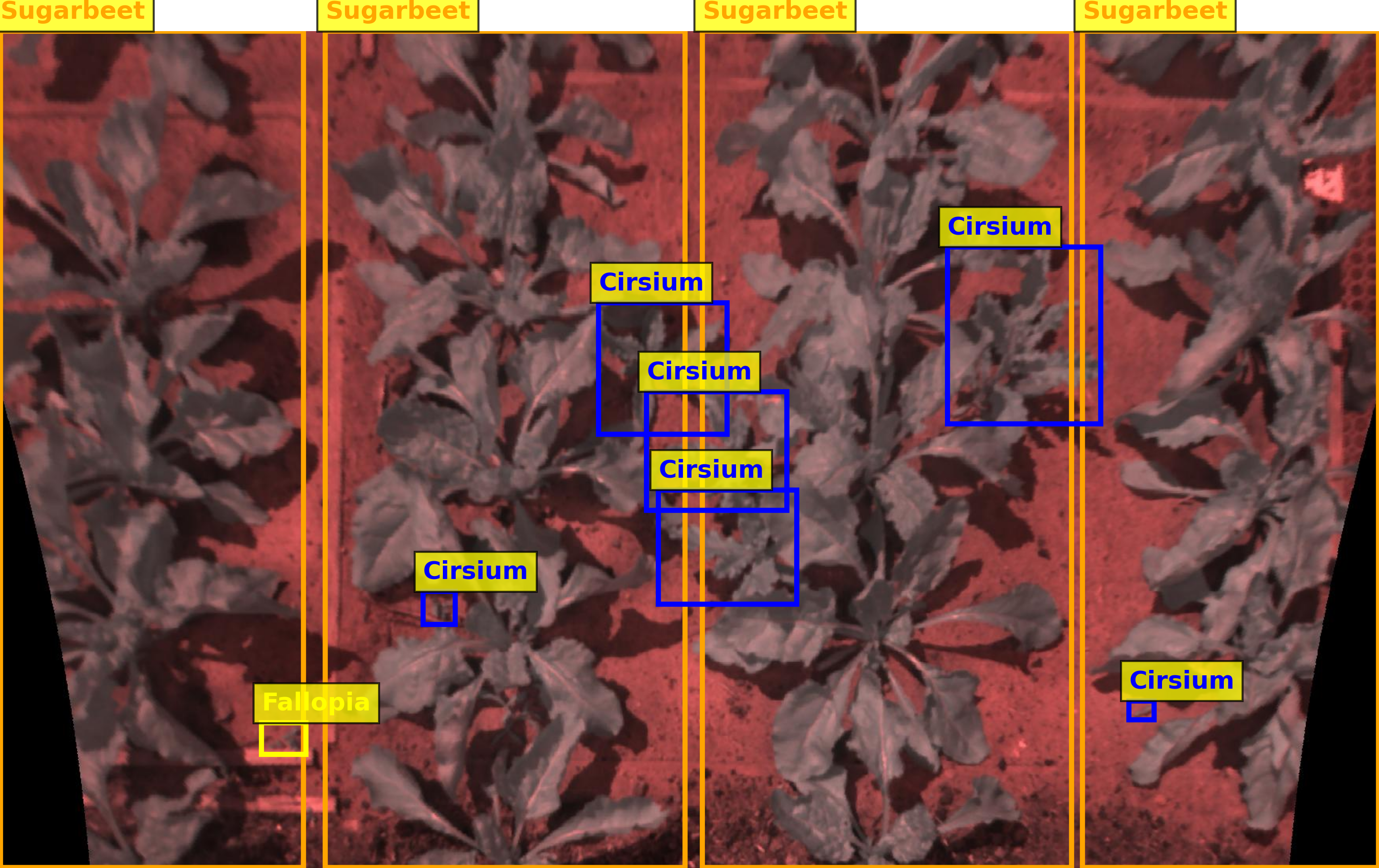}
    \caption{Annotated ground truth}
    \label{fig:annotated}
  \end{subfigure}
  \vspace{0.5em}
  \caption{Sample pseudo-RGB images from the sugar beet dataset captured in an outdoor euro-pallet setup. The left column shows the raw captured images, while the right column shows the expert-annotated ground-truth images for object detection task}
  \label{fig:comparison_groundtruth_annotated}
\end{figure}

The dataset was collected within the scope of a current project (see
Acknowledgement~\ref{subsec:ack}) using a Field Camera Unit (FCU) mounted on a smart sprayer attached to a tractor moving at 1.5m/s. Equipped with 2.3MP RGB sensors and a dual-band NIR-red filter, the system captured images at 1.1m height with a $25^\circ$ tilt. Post-processing included projection correction and pseudo-RGB generation from NIR and red channels. 
The final dataset as used in this paper consists of 2,074 images (1752×1064 px) annotated by weed scientist experts, containing sugar beet as the crop and four weed species: Cirsium, Convolvulus, Fallopia, and Echinochloa (cf. Fig.~\ref{fig:comparison_groundtruth_annotated}).

\subsection{YOLO11}
YOLO11, released by Ultralytics in September 2024, is a state-of-the-art real-time object detector optimized for low latency and high throughput on diverse hardware. It comes in five size variants—Nano (n), Small (s), Medium (m), Large (l), and Extra-large (x)—each adjusting width and depth multipliers to balance computational cost and accuracy~\cite{Jocher_Ultralytics_YOLO_2023,LearnOpenCV_YOLO11}.  

In terms of architecture, YOLO11 builds on YOLOv8 with three principal innovations. First, the traditional C2f modules in both backbone and neck are replaced by the C3k2 block: a Cross Stage Partial construction that splits a larger convolution into two smaller-kernel convolutions, reducing parameter count while preserving representational power. Second, the Spatial Pyramid Pooling—Fast (SPPF) module is retained to aggregate multi-scale context efficiently. Third, a new Cross Stage Partial with Spatial Attention (C2PSA) block is inserted immediately after SPPF to apply an area-based attention mechanism, sharpening the model’s focus on semantically important regions~\cite{Khanam2024}.  

Beyond these blocks, YOLO11 preserves an anchor-free, decoupled head for parallel classification and bounding-box regression, leveraging CIoU loss for tighter localization and label smoothing for stable convergence. Its training pipeline employs advanced data augmentations (mosaic, mixup, random flips/rotations), auto-anchor optimization, and adaptive image resizing to boost robustness. Benchmarks report consistent mAP gains over earlier YOLO versions at comparable or lower inference latency, making YOLO11 particularly well-suited for edge-device deployment in real-time systems~\cite{Khanam2024}.

\subsection{Channel-wise Knowledge Distillation}
\label{subsec:cwd}

Channel-wise Knowledge Distillation (CWD) is a feature-based approach tailored for dense prediction tasks, such as semantic segmentation and object detection, where spatial alignment methods can be noisy due to background clutter. Instead of matching features at each spatial location, CWD treats each channel’s activation map as a probability distribution over spatial positions, transferring the teacher’s “attention” to the student by minimizing a divergence between these channel-wise distributions \cite{Shu_2021_ICCV}.

Concretely, let $y^T \in \mathbb{R}^{C\times W\times H}$ and $y^S \in \mathbb{R}^{C'\times W\times H}$ be the teacher’s and student’s activation tensors at a given layer, with $C$ and $C'$ channels, respectively. We first align channel dimensions by a $1\times1$ convolution on the student if $C'\neq C$. Then for each channel $c$:
\begin{equation}
\phi(y_{c,i}) \;=\; \frac{\exp\!\bigl(y_{c,i}/T\bigr)}{\sum_{j=1}^{W\cdot H} \exp\!\bigl(y_{c,j}/T\bigr)},
\end{equation}
where $i$ indexes spatial locations and $T$ is a temperature hyperparameter controlling the sharpness of the softmax. Higher $T$ yields smoother, more distributed attention maps.

The channel-wise distillation loss is defined as the (asymmetric) Kullback–Leibler divergence~\cite{Kullback1951OnIA} between teacher and student distributions:
\begin{equation}
L_{\mathrm{CWD}}
\;=\;
T^2 \sum_{c=1}^{C}\sum_{i=1}^{W\cdot H}
\phi\bigl(y^T_{c,i}\bigr)\;
\log
\frac{\phi\bigl(y^T_{c,i}\bigr)}{\phi\bigl(y^S_{c,i}\bigr)}.
\end{equation}
The $T^2$ factor ensures appropriate gradient scaling. By focusing weight on regions where the teacher’s channel response is high, CWD naturally emphasizes foreground structures and suppresses background noise.The total training objective becomes
\begin{equation}
L = L_{\mathrm{task}} + \lambda\,L_{\mathrm{CWD}},
\end{equation}
where $L_{\mathrm{task}}$ is the standard object detection loss (classification + box regression), and $\lambda$ balances the distillation term (we set $T= 1, 2, 3, 4$ separately and $\lambda=50$ for feature maps following \cite{Shu_2021_ICCV}). 

\subsection{Masked Generative Distillation}
\label{subsec:mgd}

Masked Generative Distillation (MGD)~\cite{yang2022masked} is another feature-based distillation method that, instead of forcing the student to mimic every teacher feature directly, randomly masks a portion of the student’s feature map and trains a lightweight generative block to recover the teacher’s full features from the remaining pixels. This encourages the student to leverage their contextual information for stronger representations.

Concretely, let \(T^l\in\mathbb{R}^{C\times H\times W}\) and \(S^l\in\mathbb{R}^{C\times H\times W}\) be the teacher’s and student’s feature tensors at layer \(l\). We first sample a binary mask \(M^l\in\{0,1\}^{H\times W}\) by
\begin{equation}
M^l_{i,j} =
\begin{cases}
0, & r^l_{i,j} < \lambda,\\
1, & \text{otherwise,}
\end{cases}
\quad r^l_{i,j}\sim\mathcal{U}(0,1),
\end{equation}
where the mask ratio \(\lambda\in[0,1]\) controls the fraction of dropped pixels \cite{yang2022masked}. The masked student feature is aligned channel-wise via a \(1\times1\) convolution \(f_{\mathrm{align}}(\cdot)\), then passed through a two-layer generative projector:
\begin{equation}
G(F) \;=\; W^{(2)}\bigl(\mathrm{ReLU}(W^{(1)}F)\bigr),
\end{equation}
where \(W^{(1)},W^{(2)}\in\mathbb{R}^{C\times C\times3\times3}\) are learnable convolutional kernels \cite{yang2022masked}.

The MGD loss is defined as the squared error between the teacher’s full feature and the projector’s output on the masked student feature:
\begin{equation}
\mathcal{L}_{\mathrm{MGD}}
=
\sum_{l=1}^L
\sum_{k=1}^C
\sum_{i=1}^H
\sum_{j=1}^W
\bigl[T^l_{k,i,j} - G\bigl(f_{\mathrm{align}}(S^l)\odot M^l\bigr)_{k,i,j}\bigr]^2,
\end{equation}
where \(\odot\) denotes element-wise multiplication.

Finally, the student is trained under the combined objective
\begin{equation}
\mathcal{L}
=\;
\mathcal{L}_{\mathrm{task}}
\;+\;
\alpha\,\mathcal{L}_{\mathrm{MGD}},
\end{equation}
where \(\mathcal{L}_{\mathrm{task}}\) is the original task loss (e.g.\ cross-entropy for classification or detection/segmentation losses for dense prediction) and \(\alpha\) balances the distillation term. In our experiments we set \(\lambda=0.5\) (masking half the feature pixels each pass) and tune \(\alpha\) so that \(\mathcal{L}_{\mathrm{MGD}}\) and \(\mathcal{L}_{\mathrm{task}}\) are on comparable scales (we set $\alpha= 2\cdot10^{-5}, 4\cdot10^{-5}, 6\cdot10^{-5}, 8\cdot10^{-5}$ separately~\cite{yang2022masked}.

\subsection{Accuracy Metrics}
Detection quality is measured by the mean Average Precision at IoU=0.50 (mAP50) following the COCO evaluation protocol \cite{LinCOCO2014,Padilla2020IWSSIP}. Given a predicted box \(B_p\) and a ground-truth box \(B_{gt}\), their overlap is
\[
\mathrm{IoU}(B_p, B_{gt})
\;=\;
\frac{\lvert B_p \cap B_{gt}\rvert}{\lvert B_p \cup B_{gt}\rvert}.
\]
A prediction is a true positive if \(\mathrm{IoU}\ge0.50\); otherwise it is a false positive, and unmatched ground-truths are false negatives.

For each class \(c\), precision \(P_c\) and recall \(R_c\) at confidence threshold \(t\) are
\[
P_c(t)
=\frac{\mathrm{TP}_c(t)}{\mathrm{TP}_c(t)+\mathrm{FP}_c(t)},
\quad
R_c(t)
=\frac{\mathrm{TP}_c(t)}{\mathrm{TP}_c(t)+\mathrm{FN}_c(t)}.
\]
The Average Precision for class \(c\) at IoU=0.50 is the area under the precision–recall curve:
\[
\mathrm{AP}_{50}(c)
=\int_{0}^{1} P_c\bigl(R\bigr)\,\mathrm{d}R.
\]
Finally, mAP50 is the mean of AP\(_{50}(c)\) over all \(C\) classes:
\[
\mathrm{mAP}_{50}
=\frac{1}{C}\sum_{c=1}^{C}\mathrm{AP}_{50}(c).
\]

To complement mAP$_{50}$, we also report mAP$_{50\text{-}95}$, defined as the mean Average Precision averaged over ten IoU thresholds from 0.50 to 0.95 in increments of 0.05

\subsection{Experimental Setting}
All downstream models, namely the teacher model YOLO11x, a lightweight baseline (YOLO11n), and a distilled student (YOLO11n with CWD and MGD), were trained with a defined hyperparameter configuration (see Table~\ref{tab:hyp} along with section ~\ref{subsec:cwd} and ~\ref{subsec:mgd}). The downstream models were trained for up to 100 epochs with early stopping (patience: 20) on our sugar beet weed dataset, using five fixed random seeds per model to assess statistical robustness. We employed the AdamW optimizer~\cite{Loshchilov2019AdamW} with an initial learning rate of \(1\times10^{-3}\), weight decay of \(5\times10^{-3}\), and batch size of 32, based on preliminary hyperparameter tuning to maximize convergence speed, gradient stability, and regularization while remaining within GPU throughput limits. The dataset was split 70\%/15\%/15\% for training, validation, and testing, respectively. To solely evaluate distillation effects, automatic data augmentation was disabled for both the baseline and student models.

During distillation, we applied Channel-wise Knowledge Distillation with temperature \(T\in\{1,2,3,4\}\) and set the feature-distillation weight \(\lambda=50\) with logit-distillation weight \(\alpha=3\) in accordance with Shu et al.~\cite{Shu_2021_ICCV}. We additonally applied Masked Generative Distillation with logit-weight \(alpha\in\{2\cdot10^5, 4\cdot10^5, 6\cdot10^5, 8\cdot10^5\}\) \cite{yang2022masked}. To focus our evaluation on the detection of the target weed species, we excluded the sugar beet (crop) class from the computation of mAP$_{50}$ and mAP$_{50\text{-}95}$. Since sugar beet instances dominate the dataset and are not directly relevant to selective herbicide application, including them would artificially inflate the overall precision and average precision scores. Concretely, during per–class and mean–average–precision calculations (using the COCO evaluation protocol), we filter out all ground‐truth and predicted boxes with class label “sugar beet” and compute Precision, Recall, mAP$_{50}$ and mAP$_{50\text{-}95}$ only over the four weed classes (\textit{Cirsium}, \textit{Convolvulus}, \textit{Fallopia}, and \textit{Echinochloa}).

\begin{figure*}[!t]
  \centering
  \includegraphics[width=1\linewidth]{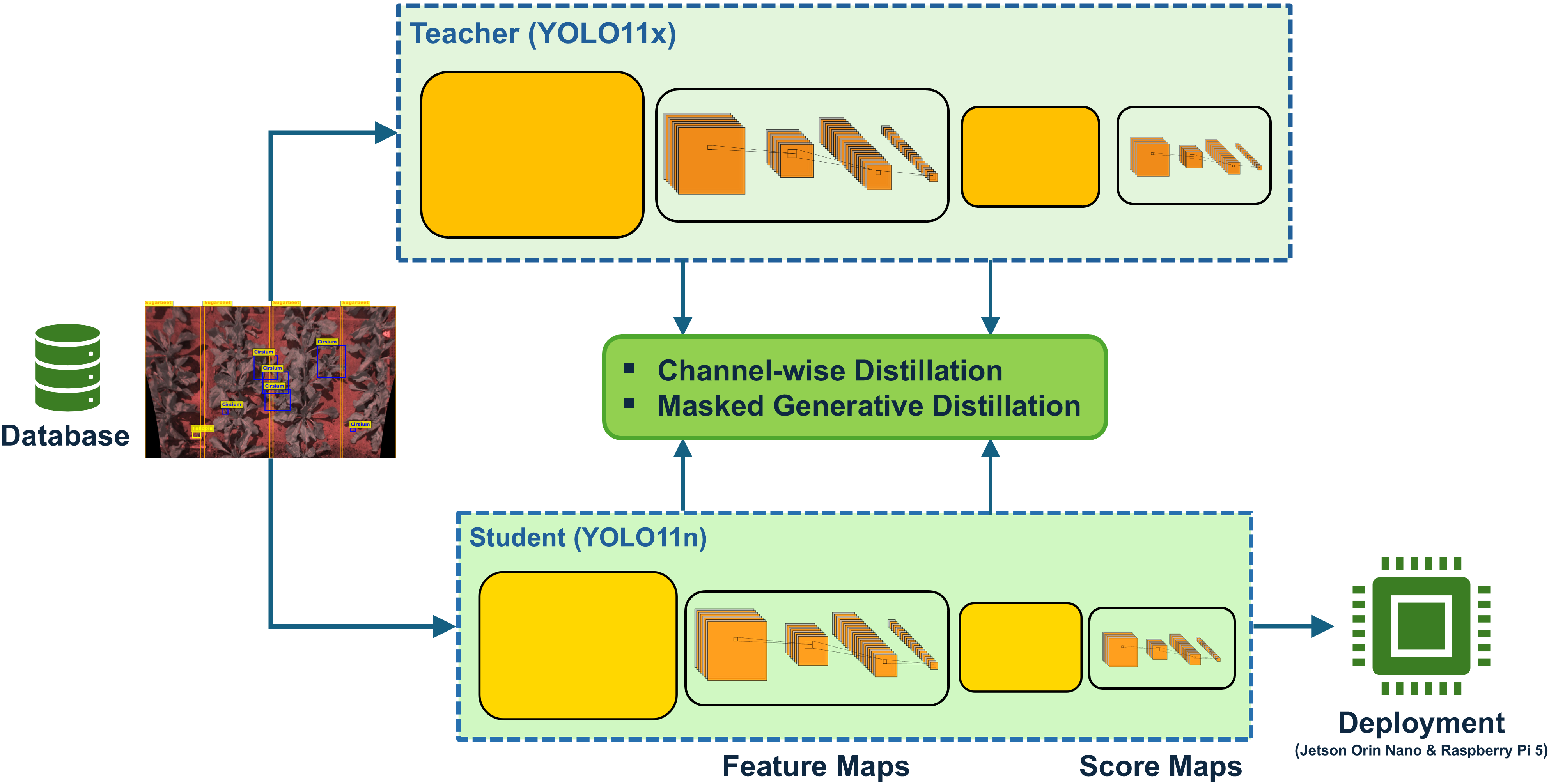}
  \caption{Overview of the knowledge distillation pipeline, transferring per-channel and class logit knowledge from the teacher YOLO11x to the student model (YOLO11n) during training by channel-wise and masked generative distillation. The resulting distilled student is then deployed on embedded devices for efficient, real-time inference.}
  \label{fig:demo}
\end{figure*}

Training was conducted on a server equipped with an NVIDIA A100-SXM4-80GB GPU, 50 GB RAM, and an AMD EPYC 75F3 32-core CPU. For real-time inference benchmarking, we deployed the distilled YOLO11n model on the NVIDIA Jetson Orin Nano (see ~\ref{subsec:Nano}) and Raspberry Pi 5 (see ~\ref{subsec:Rasp}) devices.
\subsubsection{NVIDIA Jetson Orin Nano}
\label{subsec:Nano}
The Jetson Orin Nano is equipped with an 8-core ARM Cortex-A78AE CPU operating at frequencies up to \unit[2.2]{GHz}, complemented by a 1024-core Ampere GPU featuring 32 Tensor Cores, which collectively deliver up to \unit[40]{TOPS} of mixed-precision throughput. It is designed with \unit[8]{GB} of LPDDR5 memory, offering a bandwidth of \unit[68]{GB/s}, and \unit[16]{GB} of eMMC storage, thereby facilitating expedited data staging for both batch and streaming inference processes. Hardware capabilities include support for FP16 and INT8 quantization through NVIDIA TensorRT\footnote {\url{https://github.com/NVIDIA/TensorRT} (accessed on 22 June 2025)}, in conjunction with automatic layer fusion, kernel auto-tuning, and dynamic tensor memory management. These attributes render the device particularly suitable for low-latency applications in object detection and segmentation models\footnote {\url{https://www.reichelt.com/de/en/shop/product/nvidia_jetson_orin_nano_40_tops_8_gb_ram-358840} (accessed on 22 June 2025)}.
\subsubsection{Raspberry Pi 5}
\label{subsec:Rasp}
The Raspberry Pi 5 integrates a Broadcom BCM2712 SoC with a quad-core Arm Cortex-A76 CPU (up to \unit[2.4]{GHz}) and a VideoCore VII GPU, paired with up to \unit[16]{GB} of unified LPDDR4X memory (\unit[5500]{MT/s}) and NVMe storage over a PCIe 2.0×1 interface. Instead of a dedicated NPU, it exploits NEON-accelerated FP16 arithmetic and OpenMP-enabled multi-core parallelism. Models are deployed via the NCNN\footnote{\url{https://github.com/Tencent/ncnn} (accessed on 22 June 2025)} framework, utilizing kernel fusion, packed-weight formats, and vectorized execution to achieve sustained inference rates\footnote{\url{https://www.raspberrypi.com/products/raspberry-pi-5/} (accessed on 22 June 2025)}.
\begin{table}[htbp]
  \centering
  \caption{Hyperparameter configuration for downstream model training.}
  \label{tab:hyp}
  \small

    \begin{tabular}{ll}
      \toprule
      \textbf{Hyperparameter}      & \textbf{Value}         \\ 
      \midrule
      Epochs                       & 100                    \\
      Image size                   & 640\,px                \\
      Early stopping patience      & 20                     \\
      Batch size                   & 32                     \\
      Initial learning rate        & \(1\times10^{-3}\)     \\
      Weight decay                 & \(5\times10^{-3}\)     \\
      Optimizer                    & AdamW                  \\
      Augmentation                 & Disabled               \\
      Pretrained                   & True (COCO dataset)    \\
      \bottomrule
    \end{tabular}%
  
\end{table}

\section{Results}
\label{sec:results}

\begin{figure}[!t]
  \centering
  \includegraphics[width=1\linewidth]{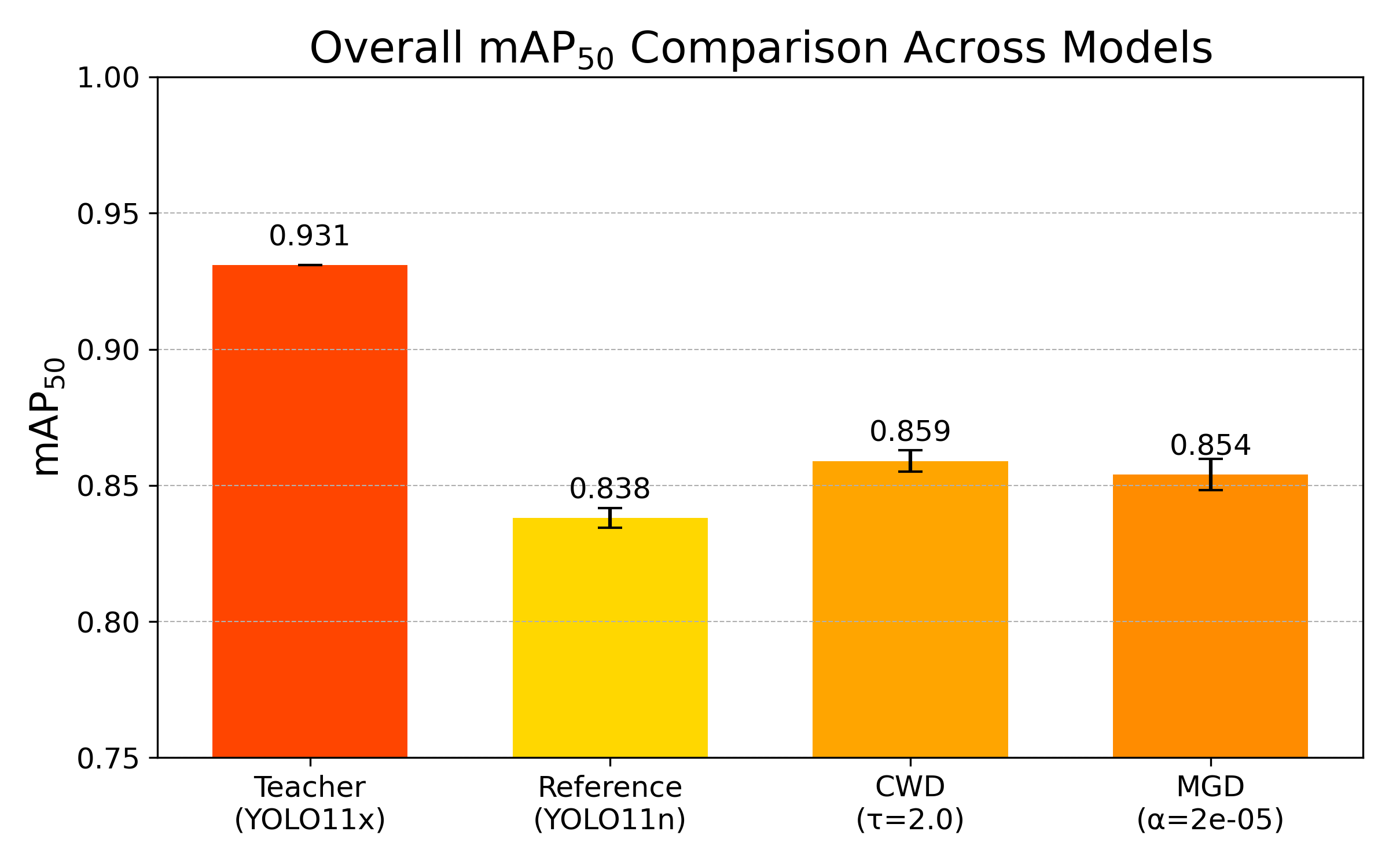}
  \caption{Overall mAP$_{50}$ comparison between the YOLO11x teacher, YOLO11n reference (mean~$\pm$~std), and distilled students using CWD ($\tau$=2.0) and MGD ($\alpha$=2\,$\times$\,10$^{-5}$).}
  \label{fig:overall_map50_comparison_four_customcolors}
\end{figure}

\begin{table}[!t]
\centering
\caption{Class-wise detection performance of the YOLO11n baseline model (mean~$\pm$~std) over the five seeds.}
\resizebox{\linewidth}{!}{
\label{tab:pure_classwise}
\begin{tabular}{l cccc}
\toprule
Class          & Precision           & Recall              & AP$_{50}$          & AP$_{50\text{-}95}$    \\
\midrule
Cirsium        & $0.914 \pm 0.035$   & $0.772 \pm 0.041$   & $0.874 \pm 0.012$   & $0.685 \pm 0.012$       \\
Convolvulus    & $0.855 \pm 0.046$   & $0.728 \pm 0.050$   & $0.825 \pm 0.015$   & $0.532 \pm 0.008$       \\
Echinochloa    & $0.789 \pm 0.038$   & $0.796 \pm 0.016$   & $0.836 \pm 0.004$   & $0.522 \pm 0.005$       \\
Fallopia       & $0.813 \pm 0.028$   & $0.755 \pm 0.028$   & $0.824 \pm 0.004$   & $0.556 \pm 0.007$       \\
\bottomrule
\end{tabular}
}
\end{table}

\subsection{Results of CWD}
\label{subsec:results_cwd}

\begin{figure}[!t]
  \centering
  \includegraphics[width=1\linewidth]{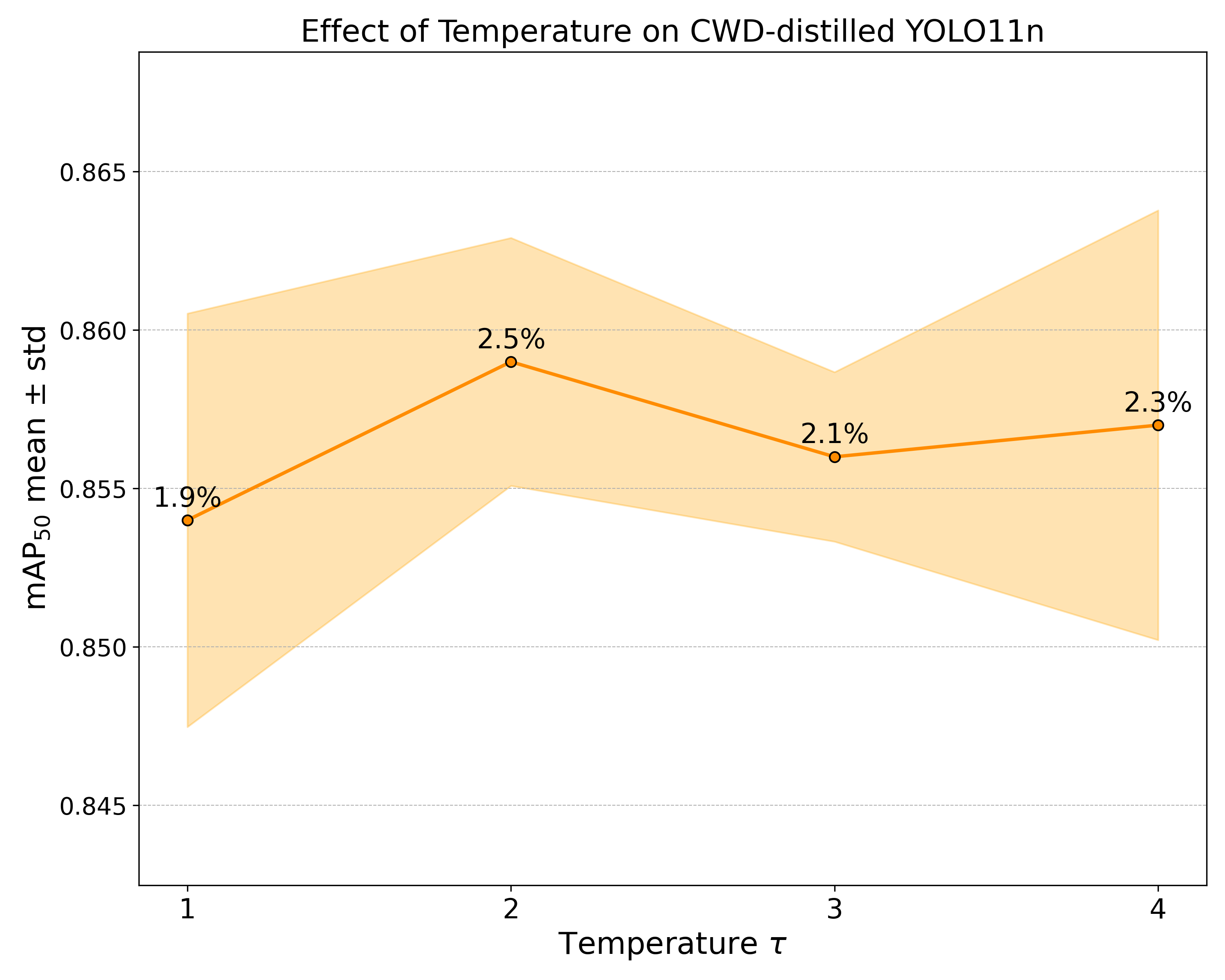}
  \caption{Effect of temperature $\tau$ on the mAP$_{50}$ performance of the CWD‐distilled YOLO11n student model. The shaded region indicates ±1 std over the five seeds, and annotations denote relative improvement over the pure baseline.}
  \label{fig:cwd_map50_vs_tau_shaded_annotated}
\end{figure}

Table~\ref{tab:cwd_summary} reports the overall detection performance of the YOLO11n student model distilled with CWD at several temperatures $\tau$. The best result is obtained at $\tau=2.0$, yielding a mean mAP$_{50}$ of $0.859\pm0.003$, which corresponds to a $2.5\%$ relative gain over the reference YOLO11n baseline and reduces the gap to the YOLO11x teacher from $7.0\%$ down to $5.3\%$ (See Table~\ref{tab:pure_overall}).  
Figure~\ref{fig:cwd_map50_vs_tau_shaded_annotated} visualizes how mAP$_{50}$ varies with $\tau$, with shaded regions indicating ±1std and annotations showing the percent improvement over the pure model.

\begin{figure}[!t]
  \centering
  \includegraphics[width=1\linewidth]{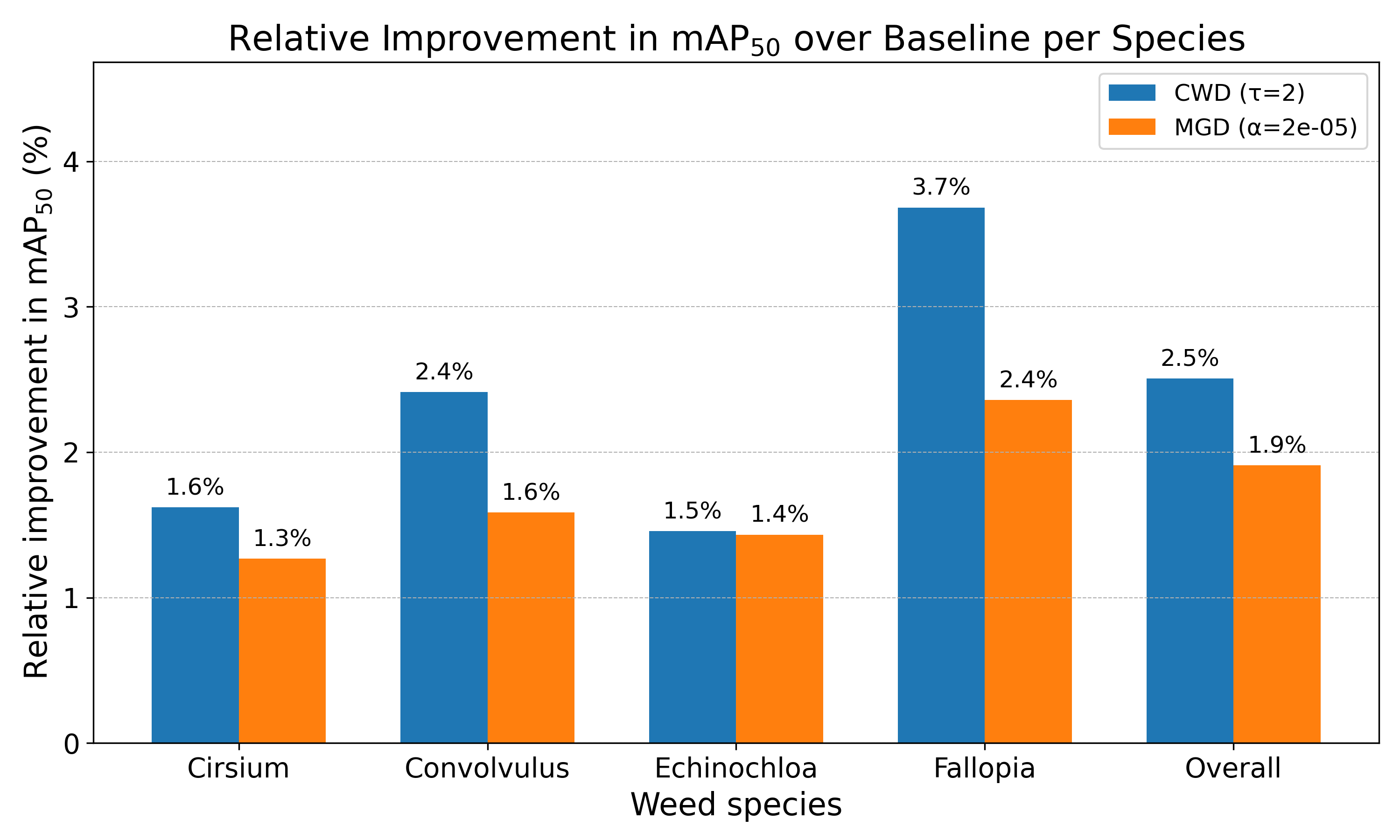}
  \caption{Relative improvement in mAP$_{50}$ of the CWD ($\tau$=2.0) and MGD ($\alpha$=2\,$\times$\,10$^{-5}$) distilled YOLO11n student models based on best parameter results over the YOLO11n reference, shown per weed species.}
  \label{fig:relative_improvement_over_classes}
\end{figure}

A per‐species breakdown is given in Table~\ref{tab:cwd_classwise}. CWD most strongly benefits the dicots \textit{Fallopia} (+3.7\%), \textit{Convolvulus} (+2.4\%) and \textit{Cirsium} (+1.6\%), but also yields consistent gains on the monocot \textit{Echinochloa} (+1.5\%). Figure~\ref{fig:relative_improvement_over_classes} plots these class‐wise improvements, highlighting CWD’s ability to transfer the teacher’s channel‐wise attention patterns and disambiguate visually similar dicot species, a critical requirement in plant phenotyping workflows.

\begin{table}[!t]
  \centering
  \small
  \caption{Overall detection performance of the CWD‐distilled YOLO11n student model at different temperatures (mean~$\pm$~std) over five seeds. The best values of each metric are highlighted in bold.}
  \resizebox{\linewidth}{!}{
  \label{tab:cwd_summary}
  \begin{tabular}{ccccc}
    \toprule
    $\tau$ & Precision & Recall & mAP$_{50}$ & mAP$_{50\text{-}95}$ \\
    \midrule
    1.0 & \textbf{0.862 $\pm$ 0.020} & 0.756 $\pm$ 0.020 & 0.854 $\pm$ 0.006 & 0.575 $\pm$ 0.009 \\
    2.0 & 0.857 $\pm$ 0.017 & 0.776 $\pm$ 0.014 & \textbf{0.859 $\pm$ 0.003} &\textbf{ 0.578 $\pm$ 0.003} \\
    3.0 & 0.856 $\pm$ 0.021 & 0.774 $\pm$ 0.019 & 0.856 $\pm$ 0.002 & 0.574 $\pm$ 0.005 \\
    4.0 & 0.847 $\pm$ 0.012 & \textbf{0.782 $\pm$ 0.021} & 0.857 $\pm$ 0.006 & 0.576 $\pm$ 0.006 \\
    \bottomrule
  \end{tabular}
  }
\end{table}

\begin{table}[!t]
  \centering
  \small
  \caption{Class‐wise detection performance of the CWD‐distilled YOLO11n student model across temperature settings (mean~$\pm$~std) over the five different seeds. The best values of each metric of each class are highlighted in bold.}
  \label{tab:cwd_classwise}
  \resizebox{\linewidth}{!}{%
    \begin{tabular}{c l c c c c}
      \toprule
      $\tau$ & Class         & Precision        & Recall             & AP$_{50}$            & AP$_{50\text{-}95}$     \\
      \midrule
      1.0    & Cirsium       & $0.909\pm0.030$              & $0.797\pm0.010$        & $0.883\pm0.012$        & $0.682\pm0.015$        \\
             & Convolvulus   & $0.871\pm0.018$              & $0.709\pm0.043$        & $0.835\pm0.015$        & $0.527\pm0.015$        \\
             & Echinochloa   & $\bm{0.832}\pm\bm{0.030}$    & $0.778\pm0.024$        & $\bm{0.855}\pm\bm{0.007}$ & $\bm{0.529}\pm\bm{0.004}$ \\
             & Fallopia      & $0.833\pm0.040$              & $0.739\pm0.039$        & $0.841\pm0.011$        & $0.561\pm0.010$        \\
      2.0    & Cirsium       & $\bm{0.911}\pm\bm{0.015}$    & $0.800\pm0.027$        & $\bm{0.888}\pm\bm{0.009}$ & $\bm{0.685}\pm\bm{0.008}$ \\
             & Convolvulus   & $0.862\pm0.054$              & $0.735\pm0.041$        & $\bm{0.845}\pm\bm{0.007}$ & $\bm{0.532}\pm\bm{0.010}$ \\
             & Echinochloa   & $0.805\pm0.013$              & $\bm{0.800}\pm\bm{0.030}$ & $0.848\pm0.009$        & $0.522\pm0.009$        \\
             & Fallopia      & $\bm{0.847}\pm\bm{0.038}$    & $\bm{0.766}\pm\bm{0.027}$ & $\bm{0.854}\pm\bm{0.004}$ & $0.556\pm0.007$        \\
      3.0    & Cirsium       & $0.902\pm0.045$              & $\bm{0.801}\pm\bm{0.040}$ & $0.885\pm0.005$        & $\bm{0.685}\pm\bm{0.011}$ \\
             & Convolvulus   & $\bm{0.872}\pm\bm{0.020}$    & $0.746\pm0.029$        & $0.840\pm0.010$        & $0.523\pm0.006$        \\
             & Echinochloa   & $0.808\pm0.030$              & $0.796\pm0.026$        & $0.852\pm0.005$        & $0.525\pm0.005$        \\
             & Fallopia      & $0.840\pm0.020$              & $0.749\pm0.008$        & $0.844\pm0.004$        & $0.562\pm0.011$        \\
      4.0    & Cirsium       & $0.888\pm0.015$              & $\bm{0.813}\pm\bm{0.010}$ & $0.886\pm0.007$        & $0.680\pm0.009$        \\
             & Convolvulus   & $0.845\pm0.009$              & $\bm{0.787}\pm\bm{0.022}$ & $\bm{0.845}\pm\bm{0.015}$ & $0.528\pm0.011$        \\
             & Echinochloa   & $0.805\pm0.025$              & $0.793\pm0.023$        & $0.846\pm0.007$        & $0.519\pm0.006$        \\
             & Fallopia      & $\bm{0.847}\pm\bm{0.036}$    & $0.732\pm0.045$        & $0.845\pm0.007$        & $\bm{0.565}\pm\bm{0.012}$ \\
      \bottomrule
    \end{tabular}%
  } 
\end{table}

\subsection{Results of MGD}
\label{subsec:results_mgd}

\begin{figure}[!t]
  \centering
  \includegraphics[width=1\linewidth]{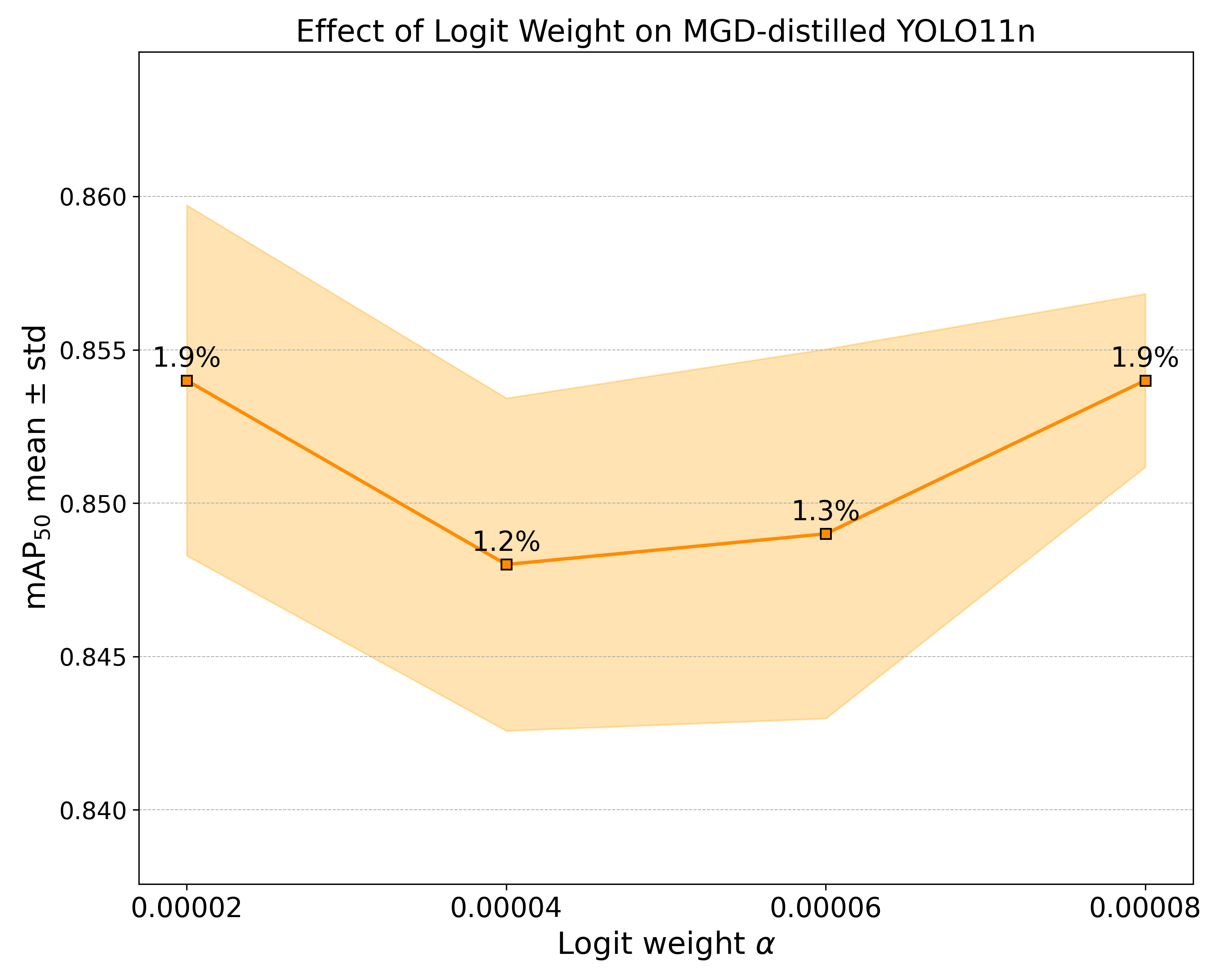}
  \caption{Effect of logit weight $\alpha$ on the mAP$_{50}$ performance of the MGD‐distilled YOLO11n student model. The shaded region indicates ±1 std over the five seeds, and annotations denote relative improvement over the pure baseline.}
  \label{fig:mgd_map50_vs_alpha_shaded_annotated}
\end{figure}

Table~\ref{tab:mgd_summary} shows the overall performance of the YOLO11n student under MGD distillation for various logit‐weight values $\alpha$. The optimal weight is $\alpha=2\times10^{-5}$, achieving mAP$_{50}=0.854\pm0.005$, or a $1.9\%$ increase over the baseline. Figure~\ref{fig:mgd_map50_vs_alpha_shaded_annotated} presents a shaded‐error‐bar plot of mAP$_{50}$ versus $\alpha$, with relative gains annotated at each point.

In Table~\ref{tab:mgd_classwise} we break down these results by species. MGD delivers its largest boost on the dicots \textit{Fallopia} (+2.4\%) and \textit{Convolvulus} (+1.6\%), but also on the monocot \textit{Echinochloa} (+1.4\%), demonstrating that logit‐based distillation can also enhance fine‐grained weed discrimination across both dicots and monocots.

\begin{table}[!t]
  \centering
  \small
  \caption{Overall detection performance of the MGD‐distilled YOLO11n student model at different logit weights (mean~$\pm$~std) over the seeds. The best values of each metric are highlighted in bold.}
  \resizebox{\linewidth}{!}{
  \label{tab:mgd_summary}
  \begin{tabular}{ccccc}
    \toprule
    $\alpha$ & Precision & Recall & mAP$_{50}$ & mAP$_{50\text{-}95}$ \\
    \midrule
    2×10$^{-5}$ & 0.850 $\pm$ 0.010 & \textbf{0.765 $\pm$ 0.007} & \textbf{0.854 $\pm$ 0.005} & 0.574 $\pm$ 0.004 \\
    4×10$^{-5}$ & 0.854 $\pm$ 0.021 & 0.758 $\pm$ 0.022 & 0.848 $\pm$ 0.005 & 0.569 $\pm$ 0.006 \\
    6×10$^{-5}$ & \textbf{0.869 $\pm$ 0.004} & 0.745 $\pm$ 0.021 & 0.849 $\pm$ 0.006 & 0.570 $\pm$ 0.008 \\
    8×10$^{-5}$ & 0.860 $\pm$ 0.017 & 0.758 $\pm$ 0.013 & 0.854 $\pm$ 0.002 &\textbf{ 0.575 $\pm$ 0.002} \\
    \bottomrule
  \end{tabular}
  }
\end{table}

\begin{table}[!t]
\centering
\small
\caption{Class‐wise detection performance of the MGD‐distilled YOLO11n student model across $\alpha$ settings (mean~$\pm$~std) over the five different seeds. The best values of each metric of each class are highlighted in bold.}
\label{tab:mgd_classwise}
\resizebox{\linewidth}{!}{
\begin{tabular}{c l c c c c}
\toprule
$\alpha$   & Class         & Precision        & Recall           & AP$_{50}$        & AP$_{50\text{-}95}$   \\
\midrule
$2\times10^5$    & Cirsium       & $0.926\pm0.016$              & $0.788\pm0.021$              & $\bm{0.885}\pm\bm{0.001}$ & $\bm{0.683}\pm\bm{0.007}$ \\
                 & Convolvulus   & $0.875\pm0.026$              & $0.699\pm0.064$              & $0.838\pm0.011$           & $0.527\pm0.011$           \\
                 & Echinochloa   & $0.793\pm0.010$              & $\bm{0.803}\pm\bm{0.014}$    & $\bm{0.848}\pm\bm{0.005}$ & $\bm{0.524}\pm\bm{0.004}$ \\
                 & Fallopia      & $0.804\pm0.031$              & $\bm{0.768}\pm\bm{0.035}$    & $0.844\pm0.009$           & $0.561\pm0.007$           \\
$4\times10^5$    & Cirsium       & $0.907\pm0.029$              & $\bm{0.795}\pm\bm{0.027}$    & $0.880\pm0.010$           & $0.678\pm0.013$           \\
                 & Convolvulus   & $0.846\pm0.035$              & $\bm{0.717}\pm\bm{0.048}$    & $0.829\pm0.014$           & $0.519\pm0.008$           \\
                 & Echinochloa   & $0.806\pm0.034$              & $0.786\pm0.030$              & $0.841\pm0.009$           & $0.522\pm0.009$           \\
                 & Fallopia      & $\bm{0.855}\pm\bm{0.016}$    & $0.732\pm0.027$              & $0.838\pm0.004$           & $0.556\pm0.007$           \\
$6\times10^5$    & Cirsium       & $\bm{0.929}\pm\bm{0.021}$    & $0.784\pm0.021$              & $0.879\pm0.011$           & $0.678\pm0.011$           \\
                 & Convolvulus   & $\bm{0.884}\pm\bm{0.036}$    & $0.667\pm0.080$              & $0.827\pm0.012$           & $0.521\pm0.011$           \\
                 & Echinochloa   & $\bm{0.815}\pm\bm{0.012}$    & $0.778\pm0.015$              & $0.841\pm0.010$           & $0.519\pm0.010$           \\
                 & Fallopia      & $0.844\pm0.054$              & $0.748\pm0.059$              & $\bm{0.847}\pm\bm{0.010}$ & $0.562\pm0.013$           \\
$8\times10^5$    & Cirsium       & $0.917\pm0.030$              & $0.778\pm0.024$              & $0.882\pm0.005$           & $0.681\pm0.006$           \\
                 & Convolvulus   & $0.879\pm0.014$              & $0.717\pm0.034$              & $\bm{0.841}\pm\bm{0.012}$ & $\bm{0.531}\pm\bm{0.013}$ \\
                 & Echinochloa   & $0.807\pm0.025$              & $0.784\pm0.023$              & $0.846\pm0.007$           & $0.519\pm0.006$           \\
                 & Fallopia      & $0.837\pm0.025$              & $0.752\pm0.035$              & $0.846\pm0.004$           & $\bm{0.567}\pm\bm{0.006}$ \\
\bottomrule
\end{tabular}
}
\end{table}

\subsection{Statistical Significance of Distillation Gains}
\label{subsec:stat_tests}

To verify that the observed mAP$_{50}$ improvements are not due to random seed variation, we performed paired Student’s $t$‐tests and Wilcoxon signed‐rank tests on the five seed runs for each distillation configuration versus the pure YOLO11n baseline. Table~\ref{tab:stat_tests} summarizes the test statistics for all four CWD temperatures $\tau$ and four MGD logit weights $\alpha$.

\begin{table}[t]
  \centering
  \scriptsize
  \caption{Statistical test results (paired $t$‐test and Wilcoxon signed‐rank) comparing distilled YOLO11n mAP$_{50}$ against the non‐distilled baseline over five random seeds.}
  \label{tab:stat_tests}
  \begin{tabular}{lccccc}
    \toprule
    \textbf{Method} & \textbf{Param} & \textbf{$t$} & \textbf{$p$ (t‐test)} & \textbf{$W$} & \textbf{$p$ (Wilcoxon)} \\
    \midrule
    CWD ($\tau$=1.0)        & 1.0     &  3.03 & $\bm{0.039}$ & 0 & 0.062 \\
    CWD ($\tau$=2.0)        & 2.0     &  6.14 & $\bm{0.004}$ & 0 & 0.062 \\
    CWD ($\tau$=3.0)        & 3.0     &  8.55 & $\bm{0.001}$ & 0 & 0.062 \\
    CWD ($\tau$=4.0)        & 4.0     &  6.34 & $\bm{0.003}$ & 0 & 0.062 \\
    \midrule
    MGD ($\alpha$=2×10$^{-5}$)  & 2×10$^{-5}$ &  5.35 & $\bm{0.006}$ & 0 & 0.062 \\
    MGD ($\alpha$=4×10$^{-5}$)  & 4×10$^{-5}$ &  2.59 & 0.061 & 1 & 0.125 \\
    MGD ($\alpha$=6×10$^{-5}$)  & 6×10$^{-5}$ &  2.23 & 0.090 & 1 & 0.125 \\
    MGD ($\alpha$=8×10$^{-5}$)  & 8×10$^{-5}$ &  6.11 & $\bm{0.004}$ & 0 & 0.062 \\
    \bottomrule
  \end{tabular}
\end{table}

All four CWD settings yield significant mAP$_{50}$ gains under the paired $t$‐test (all $p<0.05$), while the Wilcoxon signed‐rank test shows a consistent positive shift ($W=0$, $p=0.062$) that narrowly misses the 0.05 threshold due to the small sample size ($n=5$). For MGD, the paired $t$‐test indicates statistically significant improvements at $\alpha=2\times10^{-5}$ and $\alpha=1\times10^{-4}$, while the intermediate weights show non‐significant trends. The Wilcoxon test again confirms a consistent upward shift (with $p$ ranging from 0.062 to 0.125).

\subsection{Inference Times}
\label{subsec:inference_times}

Real‐time performance of distilled student models (YOLO11n) on edge platforms is summarized in Table~\ref{tab:inference_times}. On the Jetson Orin Nano GPU, the student models achieve 20.98ms/frame, on Jetson Orin Nano CPU (63.76ms) and on Raspberry Pi 5 (104.92ms) runtimes likewise satisfy typical real‐time requirements ($<$100ms)~\cite{Milioto2018, Pedoeem2018YOLOLITE}.

\begin{table}[!t]
  \centering
  \small
  \scriptsize
  \caption{Inference time (mean~$\pm$~std) over the five seeds of distilled student model (YOLO11n) for each edge device.}
    \begin{tabular}{lc}
      \toprule
      \textbf{Edge device} & \textbf{Mean ± Std (ms)} \\
      \midrule
      NVIDIA Jetson Orin Nano (GPU) & 20.98 ± 0.13 \\
      NVIDIA Jetson Orin Nano (CPU) & 63.76 ± 1.70 \\
      Raspberry Pi 5              & 104.92 ± 0.99 \\
      \bottomrule
    \end{tabular}%
  
  \label{tab:inference_times}
\end{table}

\begin{table}[!t]
\centering
\small
\caption{Per-class detection performance of the YOLO11x teacher model.}
\label{tab:teacher_per_class}
\begin{tabular}{lcccc}
\toprule
Class          & Precision & Recall & AP$_{50}$ & AP$_{50\text{-}95}$ \\
\midrule
Cirsium    & 0.917 & 0.899 & 0.953 & 0.784 \\
Convolvulus & 0.908 & 0.863 & 0.928 & 0.667 \\
Echinochloa & 0.862 & 0.870 & 0.920 & 0.649 \\
Fallopia    & 0.881 & 0.860 & 0.923 & 0.684 \\
\bottomrule
\end{tabular}
\end{table}

\begin{table}[!t]
\centering
\small
\caption{Overall detection performance of the YOLO11n baseline model (mean~$\pm$~std) over the five different seeds, along with the YOLO11x teacher model.}
\label{tab:pure_overall}
\begin{tabular}{lcccc}
\toprule
Model   & Precision & Recall & mAP$_{50}$ & mAP$_{50\text{-}95}$ \\
\midrule
\makecell{YOLO11n\\ (baseline)}  & \makecell{0.841 \\ {$\pm$0.026}}
        & \makecell{0.761 \\ {$\pm$0.028}}
        & \makecell{0.838 \\ {$\pm$0.004}}
        & \makecell{0.572 \\ {$\pm$0.005}} \\
YOLO11x & 0.892     & 0.873  & 0.931      & 0.697               \\
\bottomrule
\end{tabular}
\end{table}

\section{Discussion}
\label{sec:discussion}

Our experiments demonstrate that both Channel‐wise Knowledge Distillation (CWD) and Masked Generative Distillation (MGD) can substantially improve the accuracy of lightweight YOLO11n detectors on a fine‐grained weed detection task, without sacrificing real‐time performance.  

\paragraph{Effectiveness of CWD vs.\ MGD.}  
CWD, which aligns per‐channel attention maps via a channel‐wise KL‐divergence loss \cite{Shu_2021_ICCV}, yielded the largest overall mAP$_{50}$ gain (2.5\% at $\tau=2.0$). It was particularly effective on visually similar dicot weeds such as \textit{Fallopia} (+3.7\%) \textit{Convolvulus} (+2.4\%), suggesting that channel‐wise feature alignment helps the student focus on discriminative spatial patterns learned by the teacher. MGD, which distills high‐level logit distributions through masked feature reconstruction \cite{yang2022masked}, 
achieved a respectable 1.9\% mAP$_{50}$ improvement at $\alpha=2\times10^{-5}$ and benefited both dicots and monocots (e.g.\ +2.4\% on \textit{Fallopia}, +1.4\% on \textit{Echinochloa}), indicating that generative reconstruction can correct miscalibrated class scores.
Our hyperparameter tuning of multiple temperature values for CWD and logit‐weight settings for MGD produce optimal settings ($\tau=2.0$, $\alpha=2\times10^{-5}$) that align closely with the findings of Shu \etal~\cite{Shu_2021_ICCV} and Yang \etal~\cite{yang2022masked}, confirming consistency with the original implementations.

\paragraph{Monocot vs.\ Dicot Performance.}  
The differential gains across species underscore the value of specialized distillation. Both CWD and MGD closed more than half of the performance gap to the YOLO11x teacher on dicots, which are often visually confounded due to similar leaf shapes and textures. Importantly, monocots (e.g.\ \textit{Echinochloa}) also saw appreciable improvements (1.4–1.5\%), showing that our methods generalize across plant morphologies, which is a critical requirement in plant phenotyping where crop‐weed separation spans diverse botanical families.

\paragraph{Statistical Significance.}

A paired Student’s t-test confirms that all four CWD temperatures yield statistically significant mAP$_{50}$ gains (p$<$0.05), and two of the four MGD weights do as well. A non-parametric Wilcoxon signed-rank test shows the same directional improvements (p$\approx$0.06), which—given the small sample size—is consistent with our t-test findings.
These results confirm that our distillation methods provide reproducible, statistically meaningful improvements over the non‐distilled baseline across random seeds, particularly for the optimal CWD and MGD hyperparameters.

\paragraph{Limitations and Future Work.}  
While we observed consistent gains, both distillation techniques occasionally struggled in heavily occluded scenes or under extreme illumination, suggesting room for augmenting teacher–student alignment with attention sparsity or adversarial robustness losses. Future work could explore hybrid distillation—combining channel‐wise and generative objectives—and extend to Transformer‐based teachers (e.g.\ RT‐DETR~\cite{Zhao_2024_CVPR}) to further enrich semantic feature transfer with more seed experiments for more robust results. Later, integrating online distillation during field deployment may allow continuous adaptation to seasonal crop variations.


\section{Conclusion}
\label{sec:conclusion}

In this work, we evaluated two feature‐based knowledge distillation strategies, Channel‐wise Knowledge Distillation (CWD) and Masked Generative Distillation (MGD), to enhance the accuracy of a lightweight YOLO11n weed detector for precision agriculture. By transferring attention and semantic information from a high‐capacity YOLO11x teacher, CWD achieved up to a 2.5\% absolute mAP$_{50}$ gain ($\tau$=2.0), and MGD delivered a 1.9\% gain ($\alpha$=2\,$\times$\,10$^{-5}$), without any increase in model size or inference latency. Class‐wise analysis showed that both methods substantially improve detection of visually similar dicots (e.g.\ \textit{Fallopia} +3.7\% for CWD, +2.4\% for MGD) while also benefiting monocot species (\textit{Echinochloa} +1.4–1.5\%). Real‐time inference experiments on NVIDIA Jetson Orin Nano and Raspberry Pi 5 confirmed that the distilled models sustain 20.98ms on GPU, 63.76ms and 104.92ms on embedded CPUs, preserving the efficiency required for in‐field phenotyping and smart‐spraying. Our findings demonstrate that CWD and MGD are practical, effective ways to boost lightweight detectors in real-time applications. Future work will explore hybrid distillation combining channel‐ and spatial‐level objectives, extend to multi‐spectral and temporal data, and validate generalization on diverse agricultural datasets.

\subsubsection*{Acknowledgements.}
\label{subsec:ack}

This research was conducted within the scope of the project ``Hochleistungssensorik für smarte Pflanzenschutzbehandlung (HoPla)'' (Grant No.  13N16327), and is supported by the Federal Ministry of Education and Research (BMBF) and VDI Technology Center on the basis of a decision by the German Bundestag.

{
    \small
    \bibliographystyle{ieeenat_fullname}
    \bibliography{main}
}

\end{document}